\pdfoutput=1
\documentclass[11pt]{article}

\usepackage[final]{acl}

\usepackage{times}
\usepackage{latexsym}
\usepackage{mwe}

\usepackage[T1]{fontenc}
\usepackage[utf8]{inputenc}

\usepackage{microtype}

\usepackage{inconsolata}

\usepackage{graphicx}
\usepackage{pgfplots}
\usepackage{geometry}
\usepackage{amsmath}
\usepackage{caption}                          % For better control over captions

\usepackage{booktabs}
\usepackage{comment}
\pgfplotsset{compat=1.18}
\usepackage{xcolor}
\usepackage{fancyvrb}
\usepackage{listings}
\usepackage{hyperref}
\usepackage{multirow}
\usepackage{siunitx} 

\definecolor{myred}{RGB}{234, 103, 96}
\definecolor{myblue}{RGB}{93, 169, 222}

\title{Do LLMs Use Cultural Knowledge Without Being Told?\\A Multilingual Evaluation of Implicit Pragmatic Adaptation}

\newcommand*\samethanks[1][\value{footnote}]{\footnotemark[#1]}

\author{
 \textbf{Mehwish Nasim\textsuperscript{1,3,}}\stepcounter{footnote}\thanks{Mehwish Nasim and Sanjeevan Selvaganapathy share first authorship, with complementary leading contributions.}
 \textbf{Sanjeevan Selvaganapathy\textsuperscript{1,}}\samethanks,\\
 \textbf{Neel Ganapathi Sabhahit\textsuperscript{1}}, 
  \textbf{Marie Griesbach\textsuperscript{3}}, 
 \textbf{Pranav Bhandari\textsuperscript{1}}, 
 \textbf{Janina Lütke Stockdiek\textsuperscript{3}}, \\
 \textbf{Lennart Schäpermeier\textsuperscript{3}}, 
 \textbf{Usman Naseem\textsuperscript{2}}, 
 \textbf{Christian Grimme\textsuperscript{3}}
\\
 \textsuperscript{1}Network Analysis and Social Influence Modelling (NASIM) Lab \\
 School of Physics Maths and Computing The University of Western Australia\\
 \textsuperscript{2}School of Computing Macquarie University\\
 \textsuperscript{3}Computational Social Science \& Systems Analysis University of Münster
\\
  \small{
   \textbf{Correspondence:} {\{mehwish.nasim,sanjeevan.selvaganapathy\}}@uwa.edu.au}
 }

\begin{document}
\maketitle
% Abstract
% Abstract
\begin{abstract}
Many benchmarks show that large language models can answer direct questions about culture. We study a different question: do they also change how they speak when culture is only implied by the situation? We evaluate 60 culturally grounded conversational scenarios across five languages in three conditions: a neutral baseline (Prompt A), an explicit cultural instruction (Prompt B), and implicit situational cueing (Prompt C). We score responses on 12 pragmatic features covering deference to authority, individual-versus-group framing, and uncertainty management. We define \textbf{\emph{Pragmatic Context Sensitivity} (PCS)} as the fraction of the Prompt A$\rightarrow$B shift that reappears under Prompt A$\rightarrow$C. Across four deployed LLMs and five languages (English, German, Hindi, Nepali, Urdu), the primary stable-only PCS mean is 0.196 ($SD = 0.113$), indicating that the models recover only about one-fifth of the pragmatic shift they can produce when instructed explicitly. Transfer is strongest for authority-related cues (0.299) and weakest for individual-versus-group framing (0.120). Uncertainty-related behaviour is mixed: hedging density exhibits negative explicit gaps in all five languages, suggesting that alignment training actively suppresses the target behaviour. Because Hindi and Urdu share core grammar yet index distinct cultural communities, we use them as a natural control; a paired analysis finds no reliable baseline difference ($t = 0.96$, $p = 0.339$, $d_z = 0.06$), suggesting that models respond primarily to linguistic structure rather than to the cultural associations a language carries. We argue that multilingual cultural pragmatics is an explicit-versus-implicit deployment problem, not only a factual knowledge problem.
\end{abstract}

% Introduction
\section{Introduction}

Large language models increasingly appear in settings where getting the content right is not enough. An answer can be factually correct yet still sound socially wrong: too direct with a superior, too individualistic in a family decision, or too casual in a risk-sensitive setting. This paper asks a simple question: if a model can produce a culturally appropriate style when told explicitly, will it also produce that style when the same context is only implied by the situation?

\begin{figure}[t]
    \centering
    \includegraphics[width=0.52\textwidth]{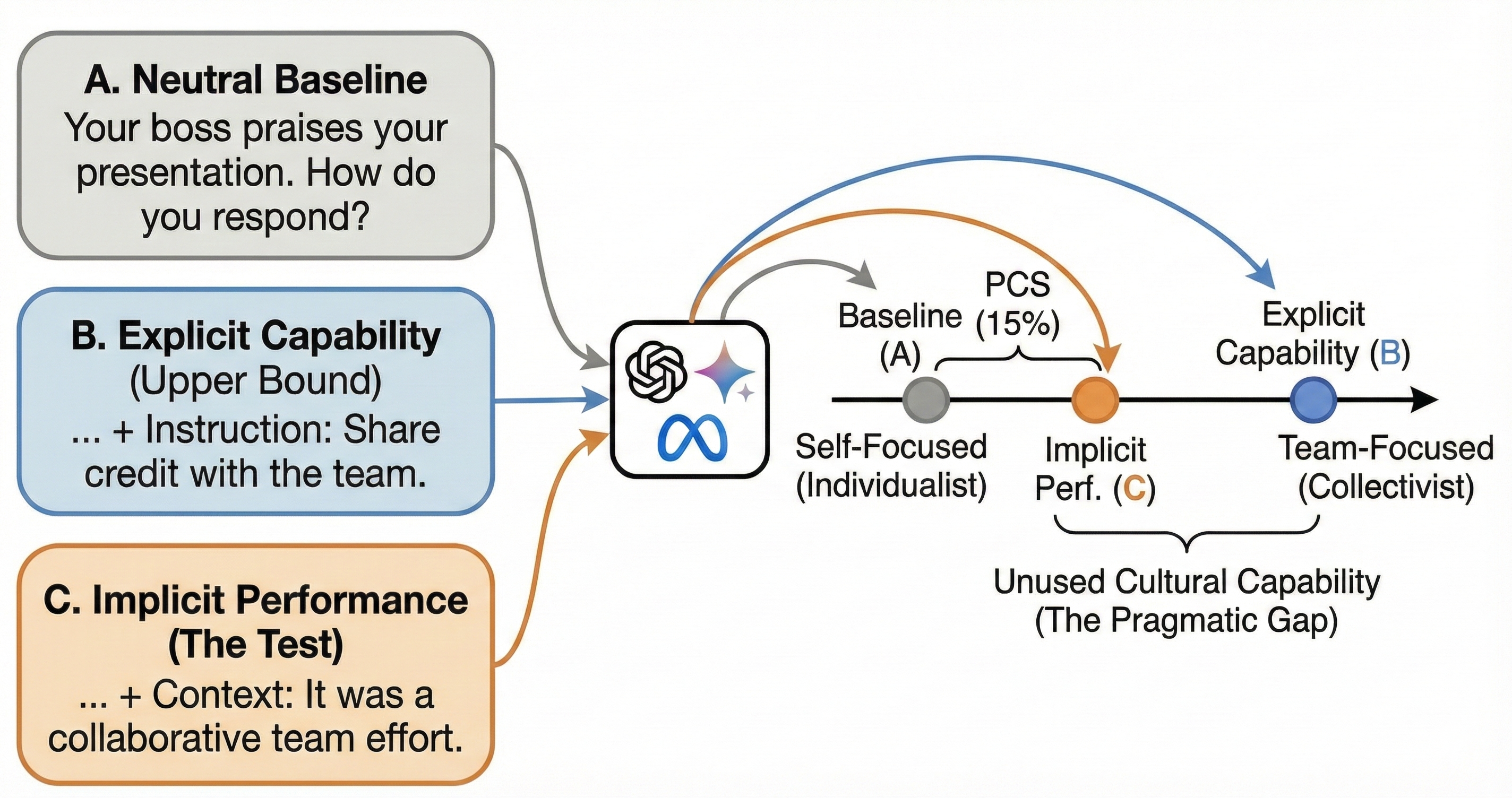}
    \caption{Three-prompt design used throughout the paper. Prompt A is the neutral baseline, Prompt B adds an explicit cultural instruction, and Prompt C adds only implicit situational cueing. PCS asks how much of the Prompt A$\rightarrow$B shift is recovered in Prompt A$\rightarrow$C.}
    \label{fig:triad_methodology}
\end{figure}

One English scenario in our dataset makes the distinction concrete. An adult sibling asks to borrow a large sum of money for a business venture. Under a neutral prompt, one model replies: ``I need a few days to look at my budget and think it over.'' Under an explicit family-duty instruction, the same model instead says: ``Of course, I will help you. We are family, and what is mine is ours.'' Under implicit cueing alone, where strong family-duty values are implied, the response moves only partway: ``I value our relationship ... but I need to be honest about the risk.'' The model can clearly produce the collectivist response style. The harder question is whether it deploys that style without being told.

We study this problem through a three-prompt design (Figure \ref{fig:triad_methodology}). Prompt~A presents a pragmatic dilemma with no extra cultural cueing. Prompt~B adds an explicit instruction describing the target cultural orientation. Prompt~C leaves out that instruction but adds situational details that should make the same pragmatic strategy appropriate for a human reader. The central quantity in the paper, \emph{Pragmatic Context Sensitivity} (PCS), asks how much of the Prompt A$\rightarrow$B shift reappears under Prompt A$\rightarrow$C.

We evaluate 60 scenarios across five languages (English, German, Hindi, Nepali, Urdu), four deployed LLMs, and three pragmatic dimensions: authority management, individual-versus-group framing, and uncertainty management. These dimensions are operational measurement axes, not claims that entire cultures can be reduced to three numbers. We adopt them because they map onto clearly scorable pragmatic choices in generated text; Table~\ref{tab:dimension_primer} introduces each dimension in plain language.

More broadly, this work addresses the problem of \emph{implicit capability deployment}. Existing benchmarks typically measure what a model can do when a task is spelled out; our setup instead asks what the model does when the relevant behaviour must be inferred from context. Cultural pragmatics serves as the testbed, but the same distinction applies to politeness, safety norms, reasoning strategies, and other behaviours that users often need without knowing how to request them explicitly.

This paper makes four contributions:
\begin{itemize}
    \item We introduce a triad evaluation design that separates baseline behaviour, explicit capability, and implicit deployment in open-ended multilingual responses.
    \item We define PCS, a simple ratio for measuring how much of an explicit behavioural shift is recovered under implicit cueing.
    \item We provide multilingual evidence that implicit transfer is limited in general and varies substantially across pragmatic dimensions.
    \item We use a Hindi-Urdu natural experiment to test whether baseline pragmatic behaviour tracks cultural association beyond shared linguistic structure.
\end{itemize}

Because the contribution is primarily methodological, we state the scope carefully. The paper does \emph{not} claim to measure all of cultural communication, nor does it treat the reported dimensions as exhaustive descriptions of any population. Instead, it provides a controlled evaluation of one question: \emph{whether current LLMs use culturally relevant pragmatic knowledge when that knowledge must be inferred rather than directly requested}.

% Related Work
\section{Related Work}

Our paper sits at the intersection of four strands of work: cultural benchmarking, explicit-versus-implicit capability gaps, pragmatic evaluation, and multilingual comparison.

\paragraph{Cultural knowledge and cultural adaptation.}
Recent work shows that LLMs often default to WEIRD (Western, Educated, Industrialised, Rich, and Democratic countries) aligned outputs and respond more strongly when cultural context is stated explicitly than when it is only implied. \citet{durmus2023globalopinionqa} found that model outputs cluster around U.S. and Western European opinion distributions unless steered toward another culture. \citet{cao2023chatgpt} reported a similar ``Americanised'' flattening effect for English prompts. More recent cultural benchmarks push beyond opinion alignment. SocialCC \citep{wu2025socialcc}, CulturalBench \citep{chiu2024culturalbench}, and NormAd \citep{rao2025normad} all show that models struggle on culturally grounded tasks even when they possess some relevant information.

\paragraph{Explicit versus implicit deployment.}
The closest conceptual precedent is the explicit--implicit localisation gap described by \citet{veselovsky2025localized}. Their setting focuses on factual cultural knowledge: models do much better when culture is named directly than when it is only implied by language choice. Our paper asks whether the same gap appears for \emph{pragmatic behaviour} in open-ended responses. In that sense, we move from ``Does the model know the fact?'' to ``Does the model change how it speaks when the situation calls for it?''

\paragraph{Pragmatic evaluation in NLP.}
Pragmatic competence has become an increasingly important evaluation target, but much of the literature focuses on English and on relatively decontextualized inference tasks. \citet{ruis2023implicature} and \citet{hu2023pragmatic} show that models often miss indirect meaning unless guided carefully. For etiquette and social norms, EtiCor \citep{dwivedi2023eticor} documents strong Western bias and weak adaptation in non-Western contexts. Our work shares that motivation, but uses a different design: instead of asking whether the model gets a pragmatic label right, we ask how much the model's open-ended response moves when contextual pressure is implicit rather than explicit.

\paragraph{Why multilingual comparison matters here.}
Multilingual benchmarks such as XTREME \citep{hu2020xtreme} and MEGA \citep{ahuja2023mega} show that model quality often drops outside English. That matters for pragmatic evaluation, but task accuracy alone cannot tell us whether communicative style changes appropriately across languages. We therefore use multilingual prompts as part of the measurement setup rather than as a mere translation stress test. Hindi and Urdu are especially informative because they share much of their grammar and core semantics while differing in script, lexicon, and cultural association \citep{prasad2012hindiurdu,nizami2020hindustani}. This makes them a useful controlled comparison for asking whether model defaults track cultural association beyond shared linguistic structure.

In summary, prior work on cultural competence in language models has largely focused on evaluating cultural knowledge, awareness, or norm sensitivity through benchmarks and classification-style tasks. 
For example, SocialCC \citep{wu2025socialcc} and CulturalBench \citep{chiu2024culturalbench} assess cultural knowledge and communication abilities across contexts but do not employ a paired explicit–implicit design, limiting their ability to quantify how behaviour transfers under different cueing conditions. 
NormAd \citep{rao2025normad} moves closer by examining cultural norm appropriateness, though primarily through judgement tasks rather than generated responses. The closest conceptual antecedent is Veselovsky et al. \citet{veselovsky2025localized}, which contrasts explicit and implicit localisation, but remains confined to factual knowledge accuracy. In contrast, our work shifts the focus from what models know to how they behave, introducing a triadic design (baseline, explicit, implicit) to measure multilingual pragmatic adaptation in open-ended generation, and quantifying how much explicitly demonstrated behaviour is recovered under implicit situational cues.

% Methodology
\section{Method}

\subsection{Three Pragmatic Dimensions}

The paper uses three cultural dimensions from the Hofstede tradition \cite{hofstede1978value}  as \emph{measurement axes} for pragmatic behaviour: Power Distance (PDI), Individualism--Collectivism (IDV), and Uncertainty Avoidance (UAI). Readers do not need prior familiarity with that literature to follow the results. Table~\ref{tab:dimension_primer} gives the practical interpretation needed for this paper.

\begin{table*}[t]
\centering
\small
\resizebox{\textwidth}{!}{%
\begin{tabular}{p{2.7cm}p{3.1cm}p{4.1cm}p{4.8cm}}
\toprule
\textbf{Dimension} & \textbf{Plain-language question} & \textbf{Example scenario} & \textbf{Scored features} \\
\midrule
PDI (Power Distance) & How much should a response defer to authority and protect face? & Disagreeing with a department head in a leadership meeting & directness, deference, face-saving, communication channel \\
\addlinespace
IDV (Individualism--Collectivism) & Should the response prioritiae personal goals or family/group obligations? & Deciding whether to lend savings to an adult sibling for a risky business venture & agency attribution, outcome framing, duty-vs.-choice, relationship priority \\
\addlinespace
UAI (Uncertainty Avoidance) & Should the response favour certainty, rules, and established expertise over ambiguity? & Choosing between a well-tested treatment and a promising but less established one & hedging density, rule reference, risk framing, expert deference \\
\bottomrule
\end{tabular}
}
\caption{Conceptual interpretation of the three pragmatic dimensions used in the study. We use them as operational axes for scoring generated text, not as exhaustive descriptions of entire cultures.}
\label{tab:dimension_primer}
\end{table*}

\subsection{Triad Evaluation Design}

Each scenario is presented in three versions.

\textbf{Prompt A (neutral baseline)} gives the dilemma without extra cultural cueing. It measures the model's default way of answering.

\textbf{Prompt B (explicit instruction)} appends a direct description of the target cultural orientation. It measures what the model can do when the desired behaviour is spelled out.

\textbf{Prompt C (implicit cueing)} leaves out the instruction but adds situational details that should make the same pragmatic strategy appropriate. It measures whether the model can infer the needed behaviour from context alone.

For example, in a Power Distance workplace scenario, Prompt A simply asks how to respond to a department head's proposed workflow. Prompt B explicitly asks for a hierarchy-sensitive, face-preserving response. Prompt C instead adds implicit cues: the department head personally developed the system, presented it to executives, and colleagues plan to show public support. A human reader can infer that direct public disagreement would be face-threatening even without any explicit cultural label. Prompt C deliberately avoids names, locations, and explicit demographic markers. This makes the task harder than many real interactions, but it isolates pragmatic inference from stereotype activation. A model that shifts because it sees an Indian name is doing something different from a model that shifts because it infers the social stakes of the situation.

\subsection{Scenario Set, Languages, and Models}

We constructed 60 scenarios covering three dimensions (PDI, IDV, UAI) and four domains (Workplace, Family, Social, Institutional). Each dimension contains 20 scenarios: five in each domain. The scenarios are controlled pragmatic dilemmas rather than attempts to reproduce the full messiness of natural conversation.

We evaluate five languages: English, German, Hindi, Nepali, and Urdu. English and German provide high-resource comparison points; Hindi and Nepali extend the evaluation to South Asian settings; and Hindi-Urdu provides a natural experiment because the two languages share much of their grammar while differing in script, register, and cultural association.

We evaluate four deployed LLMs accessed through OpenRouter: Gemini-3-flash, Grok-4.1-fast, Ministral-8B, and Mimo-V2-flash. We chose them to cover multiple providers and two price tiers---\emph{frontier} models (larger, higher-cost systems from major providers, here Gemini-3-flash and Grok-4.1-fast) and \emph{budget} models (smaller or freely available alternatives, here Ministral-8B and Mimo-V2-flash)---rather than to claim exhaustive model coverage. For each model, we collected four independent samples per prompt at temperature 0.7 and max\_tokens = 2000 between January 3--5, 2026. Model Configurations are in Appendix \ref{app:models}

This yields 14,400 prompt/sample entries in the full release ($60$ scenarios $\times 5$ languages $\times 4$ models $\times 3$ prompts $\times 4$ samples). Of these, 14,270 contain usable feature scores; 130 entries have empty score objects because all judge outputs failed parsing or validation.

All scenario files, raw responses, score files, regenerated statistics, and figure scripts are included in the public release. Appendix~\ref{app:data} lists the remaining provenance limits.

\subsection{Scoring, Human Validation, and Release Notes}

Each response is scored on the four pragmatic features associated with its dimension (listed in Table~\ref{tab:dimension_primer}). For example, a Power Distance scenario is scored on directness, deference, face-saving, and communication channel. Each feature uses a 7-point scale whose anchor points describe concrete communicative behaviours, from one end of the cultural spectrum to the other (e.g., for deference, 1 = treats all parties as equals, 7 = consistently marks hierarchical distance). We aggregate over a three-judge ensemble (Mistral Small 3.1, Gemini 2.0 Flash Lite, Qwen 2.5 72B) to reduce single-model bias. The final panel passed three narrow sanity checks: 90\% construct-validity wins on 10 Prompt B versus Prompt A comparisons, Krippendorff's $\alpha = 0.66$ on 15 shared samples, and 100\% accuracy on 3 synthetic calibration items. We use these checks as evidence of basic judge discrimination, not as exhaustive validation across the full study. Full rubric details and judge-validation materials appear in Appendix~\ref{app:scoring}.

The dataset release also includes a human validation study on a fixed 24-scenario subset. For each language, the planned design included 24 Prompt A versus Prompt C comparisons and 24 Prompt B versus Prompt C comparisons. The released finalized sheets yield 120 completed A-vs-C comparisons and 119 completed B-vs-C comparisons overall. Because the release contains one finalised sheet per language and the comparisons are clustered by scenario and language, we treat these human-validation results as descriptive rather than confirmatory.

An optional \texttt{response\_validity} meta-score is present for 11,719 of the 14,270 entries with feature scores (82.1\% coverage). Because coverage is incomplete and varies sharply across files, \texttt{response\_validity} is reported descriptively but is not used as a filter in the primary statistics.

\subsection{Metrics and Statistical Analysis}
\label{sec:metrics}

We report three main quantities.

\textbf{Language Default Index (LDI)} is the mean Prompt A score for a language-feature combination. It summarises the model's baseline pragmatic tendency when no extra cueing is provided.

\textbf{Pragmatic Context Sensitivity (PCS)} measures how much of the explicit Prompt A$\rightarrow$B shift reappears under Prompt A$\rightarrow$C. For each language $l$, dimension $d$, and feature $f$, we compute
\begin{equation}
\mathrm{PCS}^{\mathrm{raw}}_{l,d,f} = \frac{\bar{C}_{l,d,f} - \bar{A}_{l,d,f}}{\bar{B}_{l,d,f} - \bar{A}_{l,d,f}},
\end{equation}
where the bars denote means after pooling model, scenario, and sample replicates within that cell.

PCS is only interpretable when the denominator---\emph{the explicit shift from Prompt A to Prompt B}, is positive and is large enough to constitute a meaningful change. If explicit prompting fails to move the score (denominator near zero) or moves it in the wrong direction (negative denominator), the ratio becomes unstable or undefined. The primary summary we report throughout the paper is therefore \emph{stable-only PCS}, which excludes any cell whose explicit denominator $\bar{B}_{l,d,f} - \bar{A}_{l,d,f}$ is zero, negative, or smaller than 0.1. Of the 60 language-dimension-feature cells, 55 meet this stability criterion and 5 do not (all involving hedging density). We retain the raw instability-inclusive PCS mean as a secondary diagnostic.

\textbf{Hindi-Urdu Divergence (HUD)} is the absolute difference between Hindi and Urdu LDI on the same feature. It is a descriptive measure of how far the two baselines separate.

Primary inferential tests operate on matched scenario-feature means rather than pooled prompt-level observations. For cross-linguistic Prompt A comparisons, we use Friedman tests within each dimension (80 matched blocks per dimension: 20 scenarios $\times$ 4 features). For within-language Prompt A versus Prompt C differences, we use paired $t$-tests on matched scenario-feature means. For Hindi versus Urdu, we use a paired test on 240 matched Prompt A scenario-feature means.

% Results
\section{Results}

%\subsection{Models Respond Far More to Explicit Than to Implicit Cueing}
\subsection{Response to Explicit vs. Implicit Cueing}

The central empirical pattern is simple: Prompt C usually moves in the same direction as Prompt B, but much less. Under the stable-only analysis defined in Section~\ref{sec:metrics}, mean PCS is $0.196$ ($SD = 0.113$) across the 55 stable language-dimension-feature cells. The instability-inclusive raw diagnostic mean is lower, $0.152$ ($SD = 0.193$), because the five unstable cells (all hedging density) are excluded from the primary summary. In other words, the tested models recover only about one-fifth of the pragmatic shift they can produce when the desired behaviour is stated explicitly.

This gap is not confined to one model. Capability utilization over stable cells ranges from 13.6\% for Ministral-8B to 19.0\% for Gemini-3-flash (Table~\ref{tab:model_summary}). Even the strongest model remains far closer to the neutral baseline than to its own explicitly instructed behaviour.

\begin{figure}[t]
\centering
\includegraphics[width=0.7\columnwidth]{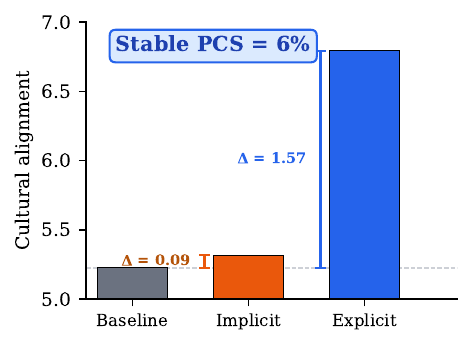}
\caption{One illustrative cell from the released results: Hindi, IDV, outcome framing. The explicit shift (Prompt A$\rightarrow$B) is large, while the implicit shift (Prompt A$\rightarrow$C) is small, yielding stable PCS $= 0.06$.}
\label{fig:competence_gap}
\end{figure}

\begin{table}[t]
\centering
\small
\resizebox{\columnwidth}{!}{
\begin{tabular}{lccccc}
\toprule
\textbf{Model} & \textbf{PCS} & \textbf{Cap.\ Util.} & \textbf{PDI} & \textbf{IDV} & \textbf{UAI} \\
\midrule
Gemini-3-flash & 0.244 & 19.0\% & 0.414 & 0.102 & 0.206 \\
Mimo-V2-flash & 0.222 & 17.3\% & 0.373 & 0.135 & 0.138 \\
Grok-4.1-fast & 0.172 & 15.3\% & 0.236 & 0.106 & 0.176 \\
Ministral-8B & 0.157 & 13.6\% & 0.219 & 0.132 & 0.111 \\
\bottomrule
\end{tabular}
}
\caption{Stable-only model summary. Mean PCS is computed over stable model-language-dimension-feature cells; capability utilisation is the percentage of the explicit shift recovered implicitly over those same stable cells.}
\label{tab:model_summary}
\end{table}

\subsection{Authority-Related Cues Transfer Best}

\begin{figure}[t]
\centering
\includegraphics[width=0.75\columnwidth]{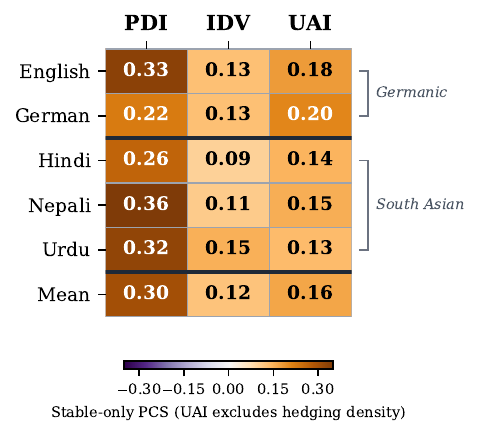}
\caption{Mean stable-only PCS by language and pragmatic dimension. Power Distance is highest across all five languages. UAI averages exclude the unstable hedging-density feature.}
\label{fig:dimension_heatmap}
\end{figure}

Transfer is not uniform across pragmatic dimensions. Stable-only mean PCS is highest for Power Distance (0.299), lowest for Individualism--Collectivism (0.120), and intermediate for Uncertainty Avoidance (0.161). Figure~\ref{fig:dimension_heatmap} shows that the Power Distance advantage appears in all five languages.

Feature-level mean PCS values tell the same story. The strongest stable features are deference (0.377) and face-saving (0.370), followed closely by communication channel. The strongest IDV feature, relationship priority, reaches only 0.186. For UAI, risk framing is the strongest stable feature at 0.210, while expert deference and rule reference remain moderate. Appendix~\ref{app:feature_pcs} provides a full ranking by median PCS.

\begin{figure}[t]
\centering
\includegraphics[width=\columnwidth]{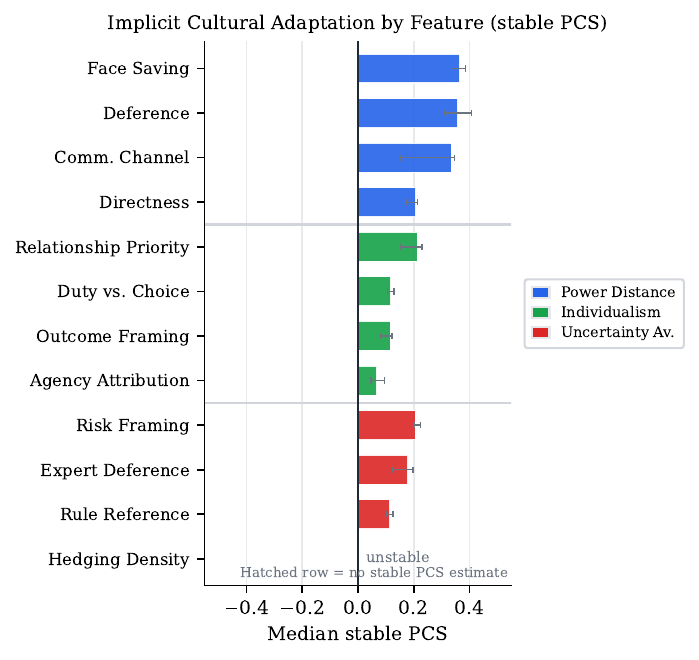}
\caption{Feature-level implicit sensitivity using median stable PCS across languages. Authority-related features dominate the top of the ranking. Hedging density is shown as unstable rather than assigned a primary PCS value.}
\label{fig:feature_pcs}
\end{figure}

The main caveat is hedging density. In all five languages, the explicit Prompt B mean for this feature is lower than the neutral Prompt A mean, so the denominator of PCS becomes negative. The raw hedging-density diagnostics are negative in every language (de $=-0.14$, en $=-0.06$, hi $=-0.38$, ne $=-0.68$, ur $=-0.40$). We therefore exclude these cells from the primary stable-only aggregate and treat them as a failure mode rather than as evidence of meaningful negative transfer.

\subsection{What Transfer Looks Like in Practice}

The qualitative examples line up with the aggregate statistics. In a Power Distance workplace scenario, a model's neutral English response already recommends a private meeting with the department head. Under explicit cueing, the response becomes overtly deferential: ``I would frame my concerns as a request for their wisdom.'' Under implicit cueing alone, the response still moves toward the same strategy: ``Never blindside a leader with negative data in a public forum.'' This kind of partial recovery is what drives the relatively high Power Distance scores.

By contrast, the weak Individualism--Collectivism transfer is visible in the family-loan scenario from the introduction. Under explicit instruction the model says, ``Of course, I will help you. We are family, and what is mine is ours.'' Under implicit cueing, however, it stops short of that collectivist shift and instead says, ``I value our relationship ... but I need to be honest about the risk.'' The implicit cues activate family sensitivity, but they do not recover the full Prompt B behaviour.

The UAI hedging-density failure mode is different again. Here the issue is not merely weak transfer but a reversal in the explicit condition: Prompt B often produces cleaner, more decisive recommendations than Prompt A, which means the explicit prompt fails to create the intended high-hedging ceiling. This makes hedging density a useful cautionary example of why denominator stability matters for PCS. Appendix~\ref{app:response_examples} provides the corresponding response excerpts in table form.

%\subsection{Hindi and Urdu Remain Close in This Controlled Setting}
\subsection{Hindi and Urdu Comparison}

The Hindi-Urdu comparison tests whether baseline pragmatic behaviour tracks cultural association beyond shared linguistic structure. In the primary paired scenario-feature analysis, Hindi and Urdu do not differ reliably: mean LDI is 5.24 for Hindi and 5.22 for Urdu ($t = 0.96$, $p = 0.339$, $d_z = 0.06$).

Feature-level differences are also small. The largest HUD occurs on expert deference (0.150), followed by relationship priority (0.130). No feature-level difference approaches a large practical separation on the 1--7 scale. Appendix~\ref{app:hindi_urdu} provides the full by-feature and by-model breakdowns, along with the Hindi-Urdu figure.

We interpret this result narrowly. It does \emph{not} show that Hindi and Urdu are pragmatically identical in general. It shows that, in this controlled prompt setup, shared linguistic structure dominates the baseline behaviour of the tested models.

%\subsection{Model Differences Exist, but They Do Not Remove the Main Gap}
\subsection{Model Differences}

Model identity is not irrelevant. Stable-only mean PCS ranges from 0.157 (Ministral-8B) to 0.244 (Gemini-3-flash), and the shared-block model comparison is statistically significant (Friedman $\chi^2 = 16.04$, $p = .0011$, Kendall's $W = 0.099$). The best-versus-worst paired contrast between Gemini and Ministral is also significant ($t = 3.82$, $p < .001$, $d_z = 0.52$).

At the same time, these differences are modest relative to the size of the overall explicit-versus-implicit gap. All four models fall far short of full implicit transfer, and both higher-cost and lower-cost systems show the same basic pattern. Four models are sufficient to establish that the gap is not an artifact of a single system, but they cannot characterise the full landscape of model-level variation; we return to this limitation in Section~\ref{sec:limitations}. Detailed model-by-language and model-by-dimension summaries appear in Appendix~\ref{app:model_results}.

%\subsection{Human Validation Shows That Many Prompt A vs. Prompt C Differences Are Subtle}
\subsection{Human Validation}

The human validation subset tempers any strong claim that Prompt C reliably produces obviously better responses than Prompt A. Across 120 completed A-vs-C comparisons, raters chose A 34 times, C 36 times, and marked 50 ties. Among decisive A-vs-C comparisons only, C receives 51.4\% of preferences, which is effectively balanced.

Across 119 completed B-vs-C comparisons, raters chose B 24 times, C 54 times, and marked 41 ties. Among decisive B-vs-C comparisons, C receives 69.2\% of preferences. Because the released study contains one finalised sheet per language and the comparisons are clustered by scenario and language, we interpret these totals descriptively rather than as confirmatory significance tests.

Taken together, the human study suggests that many of the automatically scored Prompt A versus Prompt C shifts are real but subtle. That is consistent with the main quantitative result: Prompt C does not usually behave like Prompt B, but it does not simply collapse to Prompt A either.

% Discussion
\section{Discussion}

%\subsection{The Main Empirical Claim}

% The clearest conclusion of the paper is that the tested LLMs respond much more strongly to \emph{explicit} cultural instructions than to \emph{implicit} situational cueing. This is not a claim that the models lack the relevant pragmatic behaviours altogether. Prompt B shows that they can produce those behaviours when asked directly. The gap appears in deployment: how much of that behaviour reappears when the user provides only contextual pressure.

\textbf{The Main Empirical Claim} LLMs respond far more strongly to \emph{explicit} cultural instructions than to \emph{implicit} situational cues. This does not reflect a lack of pragmatic capability. Prompt B shows they can produce such behaviours, but a deployment gap: only a fraction reappears under contextual pressure alone.

That distinction matters for NLP evaluation. Many benchmarks measure whether a model can solve a task once the desired behaviour is well specified. Our results suggest that pragmatic behaviour should also be evaluated under weaker cueing, because many real users do not know how to formulate the ``right'' cultural instruction. In that sense, the broader contribution is an evaluation perspective on implicit capability deployment.

While the aggregate pattern is shared across models, there is meaningful variation in \emph{which} dimensions each model handles best. A Friedman test over matched language--dimension--feature blocks confirms significant inter-model differences ($\chi^2 = 16.04$, $p = .001$, $W = 0.10$). Gemini-3-flash achieves the highest overall PCS (mean $= 0.244$) and displays the strongest sensitivity to Power Distance cues (PDI PCS $= 0.414$), nearly double that of Grok-4.1-fast ($0.236$) or Ministral-8B ($0.219$). Mimo-V2-flash, meanwhile, shows the highest Individualism--Collectivism adaptation (IDV PCS $= 0.135$), modestly above Ministral-8B ($0.132$) and Gemini-3-flash ($0.102$). Because all four models are closed-source, we can only speculate about the causes: Gemini's PDI advantage may reflect Google's well-documented emphasis on multilingual and multicultural training data, which could supply richer hierarchical pragmatic signal; Mimo-V2-flash's IDV edge may stem from its Chinese-language-heavy training corpus, where collectivist framings of duty and group benefit are more prevalent and thus better represented during pretraining. These hypotheses remain tentative and cannot be confirmed without access to non-public training data and fine-tuning procedures, but they highlight model-level variation in cultural competence as a key direction for future work.
%These hypotheses remain tentative---confirming them would require access to training data composition and fine-tuning procedures that are not publicly available---but they highlight model-level variation in cultural competence as an important direction for future work.

\noindent\textbf{Why Authority-Related Cues May Be Easier}
The dimension asymmetry is one of the most interpretable patterns in the paper. Power Distance cues transfer best, while Individualism--Collectivism transfers least. A plausible explanation is cue legibility. Hierarchical situations often provide multiple surface signals at once: who holds authority, whether criticism is public, whether face is at stake, and whether a private channel is available. Those signals can make the relevant pragmatic strategy relatively easy to infer.

By contrast, individual-versus-group framing often depends on subtler shifts in how a speaker balances duty, benefit, and relationship maintenance. The model may register that a family tie matters without fully reconstructing the stronger collectivist orientation seen under Prompt B. The hedging-density failure mode suggests an additional challenge for uncertainty-related behaviour: even explicit prompting does not always create the intended ceiling.

\noindent\textbf{Interpreting the Hindi Urdu Comparison}
The Hindi-Urdu comparison is useful because it holds much of the grammatical system constant while varying script, register, and cultural association. Within that controlled setting, the tested models do not show a reliable baseline difference. This points toward a simple interpretation: shared linguistic structure dominates the models' default pragmatic behaviour more than the cultural associations attached to the two registers.

This should not be overstated. The result is not a formal equivalence claim, and it does not show that Hindi and Urdu are pragmatically identical in the real world. It shows only that our prompt setup does not elicit a strong divergence between them. Still, that is informative: translating a prompt or switching scripts may not be enough to induce culturally differentiated pragmatic behaviour if the underlying linguistic system remains similar.

\noindent\textbf{The Gap Is Largest Where It Matters Most}
The practical concerns raised below are not purely speculative: a domain-level breakdown of the existing data provides direct evidence. Our 60 scenarios span four domains---Workplace, Family, Social, and Institutional, and PCS varies systematically across them (Table~\ref{tab:domain_pcs}).

\begin{table}[t]
\centering
\small
\begin{tabular}{lcccc}
\toprule
\textbf{Domain} & \textbf{PCS} & \textbf{$\bar{B}{-}\bar{A}$} & \textbf{$\bar{C}{-}\bar{A}$} & \textbf{Unrealised} \\
\midrule
Institutional & 0.155 & 1.253 & 0.210 & 1.044 \\
Family & 0.099 & 1.332 & 0.175 & 1.156 \\
Social & 0.190 & 1.375 & 0.255 & 1.120 \\
Workplace & 0.207 & 1.345 & 0.298 & 1.048 \\
\bottomrule
\end{tabular}
\caption{Scenario-level stable-only PCS by domain, pooled across all models and languages. $\bar{B}{-}\bar{A}$ is the mean explicit shift (how much the model moves when instructed), $\bar{C}{-}\bar{A}$ is the mean implicit shift (how much it moves under situational cueing alone), and Unrealised is the difference between the two---the portion of the explicit shift that implicit cueing fails to recover. Family and Institutional domains show the weakest implicit adaptation.}
\label{tab:domain_pcs}
\end{table}

Family scenarios which include sibling-loan requests and other collective-obligation dilemmas yield the lowest overall PCS (0.099), while Institutional scenarios which include questioning a doctor's recommendation and disagreeing with a professor are next lowest at 0.155. The weakest individual domain--dimension cells sharpen this pattern further: Individualism--Collectivism in Family contexts reaches only PCS $= 0.045$, meaning models recover less than 5\% of their explicit collectivist capability when family-obligation cues are implicit. Uncertainty Avoidance in Institutional contexts is also weak (PCS $= 0.121$), indicating only limited adjustment in healthcare and education settings without explicit instruction.

The domain-level pattern shows that users in culturally sensitive contexts, including patients, students, and family members navigating collective obligations, are those for whom implicit adaptation is weakest.

\noindent\textbf{Implications and Scope}
% The paper's practical implications are best read as motivated risks, now supported by the domain-level evidence above, rather than directly observed deployment failures. If a model needs an explicit instruction to adopt a less face-threatening or more group-oriented response style, users who do not know how to write that instruction may receive socially mismatched outputs. This concern is especially relevant in domains such as education, healthcare, and customer support, where users often need context-sensitive communication but may not know how to specify it.
The practical implications of this work should be understood as motivated risks rather than observed failures. If models require explicit instructions to adjust communication style, users unable to provide such prompts may receive socially mismatched outputs, particularly in domains like education, healthcare, and customer support.
We do not make causal claims. A plausible explanation is that assistant tuning enforces a default style overridden by explicit prompts but not implicit context. Our results support this hypothesis without isolating its cause and show a consistent gap across models and languages.

%We remain cautious about stronger causal claims. One plausible explanation is that current assistant tuning produces a robust default interaction style that explicit prompts can override more easily than implicit context can. Our experiments are compatible with that hypothesis, but they do not isolate the causal source of the gap. What they do show is that the problem is observable across the tested models and languages, even when the models demonstrably possess the behaviour under explicit prompting.

We note that the practical implications discussed above are inferential extrapolations from controlled probes rather than observations of deployed system behaviour. Validating whether the implicit--explicit gap documented here manifests in naturalistic conversational data where users interact freely rather than responding to structured scenarios remains an important direction for future work.

% Conclusion
\section{Conclusion}

% We asked whether LLMs use culturally relevant pragmatic knowledge when that knowledge is implied by context rather than named explicitly. Across four deployed models and five languages, the answer is: only to a limited degree. In the primary stable-only analysis, the tested models recover about one-fifth of the pragmatic shift they can produce under explicit instruction.

% The value of the paper is not only the numerical result but the evaluation framing behind it. The triad design separates default behaviour, explicit capability, and implicit deployment in a way that can be reused beyond cultural pragmatics. Our results show that open-ended multilingual behaviour can look competent under direct prompting while remaining much weaker under naturalistic cueing.

% Within the present setup, authority-related cues transfer best, individual-versus-group framing transfers least, and Hindi-Urdu baseline differences remain small. Descriptive human validation further suggests that many Prompt A versus Prompt C differences are subtle rather than dramatic. Together, these findings position multilingual cultural pragmatics as an explicit-versus-implicit deployment problem and motivate future evaluation work on behaviours that users need without knowing how to request them directly.

We show that LLMs recover only a small fraction of their pragmatic capability when cultural context is implicit rather than explicit ($\approx 20\%$ in our setting). Beyond this result, we introduce a reusable evaluation framework that separates baseline behaviour, explicit capability, and implicit deployment. Our findings highlight systematic variation across dimensions and suggest that multilingual cultural pragmatics is fundamentally an explicit-versus-implicit deployment problem.

% Limitations
\section{Limitations}
\label{sec:limitations}

\textbf{Controlled probes, not natural conversation.} The 60 scenarios are deliberately synthetic and controlled. This is a strength for isolating implicit cueing, but it also means the paper does not measure the full complexity of everyday conversation. Natural interactions contain richer discourse history, social identity cues, and turn-by-turn adaptation that our prompts intentionally suppress.

\textbf{Only 60 scenario templates.} Although the number of underlying scenario templates is 60, the factorial design (60 scenarios $\times$ 5 languages $\times$ 4 models $\times$ 3 prompts $\times$ 4 samples) yields 14{,}400 scored responses, providing substantial statistical power for the analyses we report. Established AI benchmarks such as ARC-AGI-1 use comparable or smaller evaluation sets while still supporting meaningful conclusions. Nonetheless, the paper should be read as a controlled evaluation over 60 pragmatic probes, not as exhaustive coverage of all cultural situations.

\textbf{Five languages and four models.} The study spans multiple languages and providers, but it does not justify universal claims about multilingual pragmatics. The language set is also genealogically narrow, and the model set is small relative to the rapidly changing LLM ecosystem.

\textbf{Operational framework.} We use three Hofstede-style dimensions because they map onto scorable pragmatic choices in text. We do not claim that these dimensions are exhaustive, timeless, or sufficient representations of culture. Different cultural frameworks could yield different scenario designs and different patterns.

\textbf{Aggregation sensitivity.} The paper's primary PCS summary is stable-only: cells with zero, negative, or near-zero Prompt A$\rightarrow$B gaps are excluded from the headline aggregate. This is a principled choice, but it means aggregate PCS depends on denominator handling. We therefore report the raw instability-inclusive PCS mean alongside the stable-only result.

\textbf{Judge validation scope.} The judge ensemble passes useful sanity checks, but those checks are small. They support basic discrimination, not exhaustive validity across all languages and scenario types. In addition, the public release does not include the raw GPT-4o reference scores or all raw candidate-judge trial artifacts used during panel selection.

\textbf{Human validation scope.} The human validation study is informative but limited. The released bundle contains one finalised sheet per language, and the comparisons are clustered by scenario and language. For that reason, we treat the human validation results as descriptive rather than confirmatory.

\textbf{Optional response validity.} The released score files include an optional \texttt{response\_validity} field for only 82.1\% of entries with usable feature scores. Because this field is incomplete and highly uneven across files, we do not use it as a primary inclusion criterion.

% Ethical Considerations
\section{Ethical Considerations}

\textbf{Risk of stereotyping.} The paper measures whether models shift along predefined pragmatic axes. Those axes are useful for evaluation, but they can be misread as prescriptions for how members of a language community ``should'' speak. We explicitly reject that interpretation. The dimensions are group-level abstractions used to construct controlled probes, not rules about individuals.

\textbf{Researcher interpretation.} Scenario design and rubric construction inevitably involve judgement. Although native-speaker collaborators reviewed translations and cultural appropriateness, the operationalisation still reflects researcher choices about which pragmatic behaviours to score and how to describe them.

\textbf{Human validation participants.} The main response corpus contains only model outputs generated from synthetic scenarios, but the validation component did involve human raters. We report only aggregate results and de-identified study materials in the release.

\textbf{Potential downstream misuse.} Better measurement of culturally adaptive behaviour could support more inclusive systems, but it could also be used to tailor persuasive or manipulative content more effectively. We believe documenting the current limitations is still worthwhile, because responsible deployment requires understanding what current systems do poorly.

\textbf{Use of AI tools.} AI-based language tools were used for wording and copy-editing support. The authors take responsibility for the analysis, interpretation, and final manuscript text.

\section*{Acknowledgement}
Dr Mehwish Nasim acknowledges Women In Research (WiRe) Fellowship at University of M\"{u}nster and Research Collaboration Grant (2026) at The University of Western Australia.
\bibliography{custom}

\appendix

% Appendix for the ACL manuscript

\section{Model Configuration}
\label{app:models}

Table~\ref{tab:model_config} provides full model identifiers and API configuration used for all experiments.

\begin{table*}[h]
\centering
\small
\begin{tabular}{lll}
\toprule
\textbf{Model} & \textbf{OpenRouter ID} & \textbf{Deployment tier} \\
\midrule
Grok-4.1-fast & \texttt{x-ai/grok-4.1-fast} & Premium \\
Gemini-3-flash-preview & \texttt{google/gemini-3-flash-preview} & Premium \\
Ministral-8B-2512 & \texttt{mistralai/ministral-8b-2512} & Lower-cost \\
Mimo-V2-flash & \texttt{xiaomi/mimo-v2-flash:free} & Lower-cost \\
\bottomrule
\end{tabular}
\caption{Evaluated models with OpenRouter API identifiers.}
\label{tab:model_config}
\end{table*}

\paragraph{API Parameters.} All models were queried with temperature = 0.7, max\_tokens = 2000, and 4 samples per prompt condition. Requests were made via the OpenRouter API between January 3--5, 2026.

\paragraph{System Prompt.} The following system prompt was used for all response generation:

\begin{quote}
\textit{You are a helpful assistant. Respond naturally and thoughtfully. Your response must be in \{language\_name\} only.}
\end{quote}

The \texttt{\{language\_name\}} placeholder was substituted with the target language (English, German, Hindi, Nepali, or Urdu).

\section{Scoring Rubrics and Judge Validation}
\label{app:scoring}

Responses were scored using an LLM-as-judge methodology with explicit feature-level rubrics. Each response was evaluated on four pragmatic features appropriate to its scenario dimension, yielding 12 features total: directness, deference, face-saving, and communication channel preference for Power Distance; agency attribution, outcome framing, duty-versus-choice orientation, and relationship priority for Individualism-Collectivism; and hedging density, rule reference frequency, risk framing, and expert deference for Uncertainty Avoidance. Each feature was scored on a 7-point Likert scale with explicit behavioral anchors (Tables~\ref{tab:pdi_features}--\ref{tab:uai_features}).

\subsection{Judge Ensemble}

We employed a three-judge ensemble selected for organizational diversity and validated against GPT-4o reference scores:

\begin{table}[h]
\centering
\small
\resizebox{\columnwidth}{!}{%
\begin{tabular}{llc}
\toprule
\textbf{Judge Model} & \textbf{Organization} & \textbf{$r$ (vs GPT-4o)} \\
\midrule
Mistral Small 3.1 24B & Mistral AI (France) & 0.81 \\
Gemini 2.0 Flash Lite & Google (US) & 0.82 \\
Qwen 2.5 72B & Alibaba (China) & 0.77 \\
\bottomrule
\end{tabular}
}
\caption{Judge model correlations with GPT-4o reference scores on stratified validation sample ($n = 40$).}
\label{tab:judge_validation}
\end{table}

\paragraph{Validation Protocol.} The final panel was validated on three tests:
\begin{itemize}
    \item \textbf{Construct validity}: 90\% of Prompt B responses scored higher than corresponding Prompt A responses (expected direction).
    \item \textbf{Inter-rater agreement}: Krippendorff's $\alpha = 0.66$ across the three-judge ensemble.
    \item \textbf{Synthetic calibration}: 100\% accuracy on contrastive pairs with known ground truth (artificially constructed high/low exemplars).
\end{itemize}

We evaluated four panel configurations before selecting the final ensemble:

\textbf{Configuration A (Benchmark-Optimized):} Models selected for LMArena Elo and MMLU-Pro performance (Gemini 3 Flash, Mistral Medium 3.1, Qwen3 14B). Failed all validation tests: Gemini 3 Flash showed inverted construct validity (selected A over B in 80\% of cases), Qwen3 14B exhibited ceiling effects (12/15 responses scored 7/7), and synthetic calibration accuracy was 33\%.

\textbf{Configuration B (Premium Panel):} Claude Sonnet 4, GPT-4o, Gemini 2.0 Flash. Passed construct validity (90\% B wins) and inter-rater agreement ($\alpha = 0.734$), but failed synthetic calibration on the IDV dimension (67\% accuracy).

\textbf{Configuration C (Lower-cost + DeepSeek):} Mistral Small 3.1, Gemini 2.0 Flash Lite, DeepSeek V3.2. Passed construct validity (80\% B wins) but failed inter-rater agreement ($\alpha = 0.326$) due to DeepSeek's extreme score polarization (predominantly 4s or 7s).

\textbf{Configuration D (Final):} Mistral Small 3.1, Gemini 2.0 Flash Lite, Qwen 2.5 72B. Passed all three tests: construct validity (90\% B wins), inter-rater agreement ($\alpha = 0.66$), and synthetic calibration (100\% accuracy). This configuration balances organizational diversity (France, US, China) with cost efficiency.

These iterations demonstrate that benchmark performance does not predict pragmatic evaluation capability: Configuration A's models ranked highly on LMArena but failed to discriminate culturally adapted responses.

The public release includes the final sanity-check summaries and a mixed-provenance candidate-judge selection record. It does not include the raw GPT-4o reference scores or the raw candidate-judge trial outputs used during panel selection.

\paragraph{Score Aggregation.} The release contains 14,400 scored prompt/sample entries. Of these, 130 contain no usable feature scores because all judge outputs failed parsing or validation, leaving 14,270 entries with feature scores and 57,080 feature-level observations. Final feature scores are computed as the mean across judges; higher-level summaries then aggregate across the four samples per prompt condition. An optional response\_validity meta-score is present for 11,719 of the 14,270 feature-scored entries (82.1\% coverage) and is reported descriptively rather than used as a filter.

\subsection{Feature Definitions}

Tables~\ref{tab:pdi_features}--\ref{tab:uai_features} provide the complete scoring rubrics for all 12 pragmatic features.

\begin{table*}[t]
\centering
\small
\begin{tabular}{p{2.8cm}p{4cm}p{4cm}p{4cm}}
\toprule
\textbf{Feature} & \textbf{1 (Low)} & \textbf{4 (Mid)} & \textbf{7 (High)} \\
\midrule
Directness & Blunt, direct challenge or contradiction; imperative tone & Balanced; raises concerns with some hedging & Highly indirect; uses suggestions, questions, or third-party framing \\
\addlinespace
Deference & Treats authority as equal or challenges their expertise & Neutral acknowledgement of hierarchy & Explicit respect markers; attributes wisdom/authority to superior \\
\addlinespace
Face-saving & No concern for face; public criticism acceptable & Some face-saving; suggests private discussion & Elaborate face-saving; protects superior's dignity as primary concern \\
\addlinespace
Communication channel & Advocates public, direct confrontation & Mixed approach; some private, some public elements & Strongly prefers private, one-on-one communication \\
\bottomrule
\end{tabular}
\caption{Power Distance (PDI) feature scoring anchors. Higher scores indicate more hierarchical, face-conscious communication.}
\label{tab:pdi_features}
\end{table*}

\begin{table*}[t]
\centering
\small
\begin{tabular}{p{2.8cm}p{4cm}p{4cm}p{4cm}}
\toprule
\textbf{Feature} & \textbf{1 (Individualist)} & \textbf{4 (Mid)} & \textbf{7 (Collectivist)} \\
\midrule
Agency attribution & Individual as sole agent; personal responsibility emphasized & Balanced individual and group agency & Collective/group agency; shared responsibility; family/team framing \\
\addlinespace
Outcome framing & Focus on personal benefits, self-actualization, individual goals & Balanced personal and group considerations & Focus on group harmony, family welfare, collective benefit \\
\addlinespace
Duty vs. choice & Emphasis on personal choice, autonomy, self-determination & Balanced duty and choice considerations & Emphasis on duty, obligation, role-based expectations \\
\addlinespace
Relationship priority & Task/goal completion prioritized over relationships & Balanced task and relationship concerns & Relationship maintenance prioritized; harmony over efficiency \\
\bottomrule
\end{tabular}
\caption{Individualism-Collectivism (IDV) feature scoring anchors. Higher scores indicate more collectivist framing.}
\label{tab:idv_features}
\end{table*}

\begin{table*}[t]
\centering
\small
\begin{tabular}{p{2.8cm}p{4cm}p{4cm}p{4cm}}
\toprule
\textbf{Feature} & \textbf{1 (Low UAI)} & \textbf{4 (Mid)} & \textbf{7 (High UAI)} \\
\midrule
Hedging density & Confident, unqualified assertions; minimal hedging & Moderate hedging; some qualifiers & Dense hedging; frequent ``might,'' ``could,'' ``perhaps,'' epistemic markers \\
\addlinespace
Rule reference & Appeals to flexibility, context-dependence, personal judgement & Balanced rule and flexibility references & Strong appeals to rules, policies, precedent, tradition \\
\addlinespace
Risk framing & Opportunity-focused; embraces uncertainty as potential & Balanced risk and opportunity framing & Risk-averse; worst-case thinking; threat-focused \\
\addlinespace
Expert deference & Encourages independent judgement; questioning experts acceptable & Balanced expert and personal judgement & Strong deference to experts, authorities, established wisdom \\
\bottomrule
\end{tabular}
\caption{Uncertainty Avoidance (UAI) feature scoring anchors. Higher scores indicate more uncertainty-avoiding communication.}
\label{tab:uai_features}
\end{table*}

\section{Model-by-Model Results}
\label{app:model_results}

\subsection{PCS by Model and Dimension}

Table~\ref{tab:pcs_model_dimension} reports mean Pragmatic Context Sensitivity scores broken down by model and cultural dimension.

\begin{table}[h]
\centering
\small
\begin{tabular}{llcc}
\toprule
\textbf{Model} & \textbf{Dimension} & \textbf{Mean PCS} & \textbf{SD} \\
\midrule
\multirow{3}{*}{Gemini-3-flash} & PDI & 0.414 & 0.124 \\
& IDV & 0.102 & 0.081 \\
& UAI & 0.206 & 0.076 \\
\addlinespace
\multirow{3}{*}{Grok-4.1-fast} & PDI & 0.236 & 0.129 \\
& IDV & 0.106 & 0.056 \\
& UAI & 0.176 & 0.069 \\
\addlinespace
\multirow{3}{*}{Mimo-V2-flash} & PDI & 0.373 & 0.148 \\
& IDV & 0.135 & 0.088 \\
& UAI & 0.138 & 0.086 \\
\addlinespace
\multirow{3}{*}{Ministral-8B} & PDI & 0.219 & 0.175 \\
& IDV & 0.132 & 0.069 \\
& UAI & 0.111 & 0.083 \\
\bottomrule
\end{tabular}
\caption{Stable-only PCS by model and cultural dimension. UAI rows exclude the unstable hedging-density ratios for all models, and Ministral's PDI mean excludes one unstable communication-channel cell.}
\label{tab:pcs_model_dimension}
\end{table}

\subsection{PCS by Model and Language}

Table~\ref{tab:pcs_model_language} reports mean PCS scores for each model-language combination.

\begin{table}[h]
\centering
\small
\begin{tabular}{llcc}
\toprule
\textbf{Model} & \textbf{Language} & \textbf{Mean PCS} & \textbf{SD} \\
\midrule
\multirow{5}{*}{Gemini-3-flash}
& German & 0.237 & 0.151 \\
& English & 0.294 & 0.169 \\
& Hindi & 0.215 & 0.169 \\
& Nepali & 0.266 & 0.226 \\
& Urdu & 0.206 & 0.114 \\
\addlinespace
\multirow{5}{*}{Grok-4.1-fast}
& German & 0.172 & 0.121 \\
& English & 0.203 & 0.122 \\
& Hindi & 0.143 & 0.093 \\
& Nepali & 0.162 & 0.106 \\
& Urdu & 0.182 & 0.098 \\
\addlinespace
\multirow{5}{*}{Mimo-V2-flash}
& German & 0.179 & 0.115 \\
& English & 0.221 & 0.164 \\
& Hindi & 0.195 & 0.121 \\
& Nepali & 0.269 & 0.257 \\
& Urdu & 0.246 & 0.107 \\
\addlinespace
\multirow{5}{*}{Ministral-8B}
& German & 0.149 & 0.065 \\
& English & 0.126 & 0.070 \\
& Hindi & 0.110 & 0.098 \\
& Nepali & 0.215 & 0.186 \\
& Urdu & 0.187 & 0.166 \\
\bottomrule
\end{tabular}
\caption{Stable-only PCS by model and language. Most rows summarize 11 stable features; Ministral-Nepali summarizes 10 because two cells are unstable.}
\label{tab:pcs_model_language}
\end{table}

\subsection{Model Summary Statistics}

Table~\ref{tab:model_aggregate_summary} provides aggregate statistics for each evaluated model.

\begin{table}[h]
\centering
\small
\resizebox{\columnwidth}{!}{%res
\begin{tabular}{lccccc}
\toprule
\textbf{Model} & \textbf{PCS} & \textbf{Cap. Util.} & \textbf{$\bar{A}$} & \textbf{$\bar{B}$} & \textbf{$\bar{C}$} \\
\midrule
Gemini-3-flash & 0.244 & 19.0\% & 5.24 & 6.76 & 5.53 \\
Mimo-V2-flash & 0.222 & 17.3\% & 5.17 & 6.61 & 5.42 \\
Grok-4.1-fast & 0.172 & 15.3\% & 5.00 & 6.67 & 5.26 \\
Ministral-8B & 0.157 & 13.6\% & 5.03 & 6.38 & 5.21 \\
\bottomrule
\end{tabular}
}
\caption{Stable-only model summary. $\bar{A}$, $\bar{B}$, and $\bar{C}$ are mean scores over stable model-language-dimension-feature cells. Capability utilization is the percentage of $\Delta_{AB}$ captured by $\Delta_{AC}$ over those same stable cells.}
\label{tab:model_aggregate_summary}
\end{table}

\section{Complete LDI Scores}
\label{app:ldi}

\begin{figure}[t]
\centering
\includegraphics[width=\columnwidth]{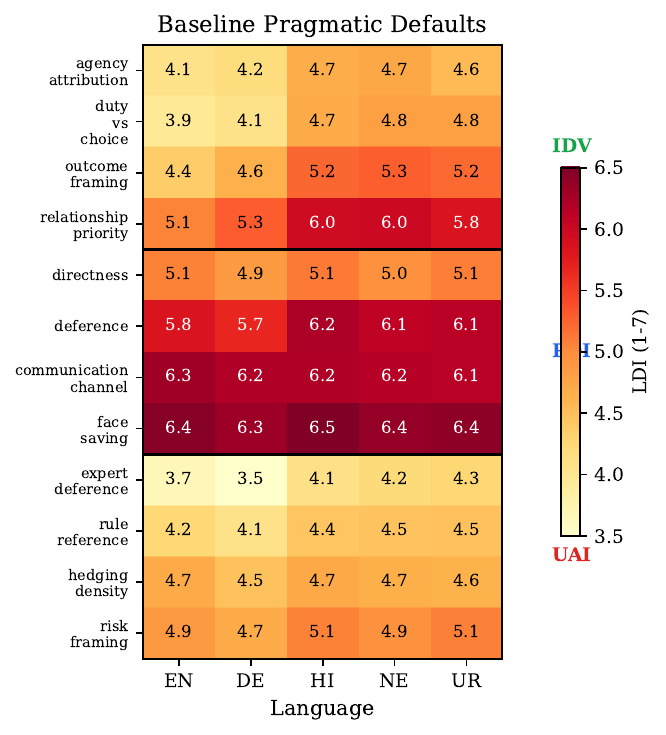}
\caption{Language Default Index (LDI) heatmap across all 12 features and five languages. Cell values are Prompt A means on the 1--7 scale, and darker cells indicate higher baseline scores. Rows are grouped by dimension.}
\label{fig:ldi_heatmap}
\end{figure}

Table~\ref{tab:ldi_full} reports Language Default Index scores for all language-feature combinations. LDI represents the mean score on Prompt A (neutral baseline) responses. Figure~\ref{fig:ldi_heatmap} provides a heatmap view of the same baseline pattern.

\begin{table*}[t]
\centering
\small
\begin{tabular}{llccccc}
\toprule
\textbf{Dimension} & \textbf{Feature} & \textbf{German} & \textbf{English} & \textbf{Hindi} & \textbf{Nepali} & \textbf{Urdu} \\
\midrule
\multirow{4}{*}{IDV}
& Agency attribution & 4.17 $\pm$ 0.88 & 4.08 $\pm$ 0.80 & 4.67 $\pm$ 0.93 & 4.73 $\pm$ 0.94 & 4.55 $\pm$ 0.82 \\
& Duty vs. choice & 4.07 $\pm$ 0.73 & 3.95 $\pm$ 0.67 & 4.68 $\pm$ 0.71 & 4.81 $\pm$ 0.77 & 4.81 $\pm$ 0.75 \\
& Outcome framing & 4.65 $\pm$ 0.93 & 4.37 $\pm$ 0.85 & 5.23 $\pm$ 0.90 & 5.28 $\pm$ 0.95 & 5.20 $\pm$ 0.84 \\
& Relationship priority & 5.29 $\pm$ 0.89 & 5.05 $\pm$ 0.87 & 5.95 $\pm$ 0.76 & 5.98 $\pm$ 0.83 & 5.82 $\pm$ 0.80 \\
\addlinespace
\multirow{4}{*}{PDI}
& Communication channel & 6.18 $\pm$ 0.91 & 6.29 $\pm$ 0.76 & 6.18 $\pm$ 0.85 & 6.15 $\pm$ 0.88 & 6.15 $\pm$ 0.88 \\
& Deference & 5.66 $\pm$ 0.71 & 5.82 $\pm$ 0.77 & 6.22 $\pm$ 0.62 & 6.08 $\pm$ 0.73 & 6.13 $\pm$ 0.67 \\
& Directness & 4.86 $\pm$ 0.72 & 5.06 $\pm$ 0.87 & 5.12 $\pm$ 0.88 & 4.95 $\pm$ 0.92 & 5.08 $\pm$ 0.87 \\
& Face-saving & 6.31 $\pm$ 0.46 & 6.44 $\pm$ 0.44 & 6.49 $\pm$ 0.45 & 6.36 $\pm$ 0.57 & 6.40 $\pm$ 0.49 \\
\addlinespace
\multirow{4}{*}{UAI}
& Expert deference & 3.47 $\pm$ 1.55 & 3.65 $\pm$ 1.47 & 4.11 $\pm$ 1.40 & 4.21 $\pm$ 1.41 & 4.26 $\pm$ 1.38 \\
& Hedging density & 4.47 $\pm$ 0.74 & 4.69 $\pm$ 0.71 & 4.71 $\pm$ 0.81 & 4.66 $\pm$ 0.83 & 4.62 $\pm$ 0.80 \\
& Risk framing & 4.68 $\pm$ 1.34 & 4.88 $\pm$ 1.25 & 5.06 $\pm$ 1.16 & 4.94 $\pm$ 1.30 & 5.08 $\pm$ 1.17 \\
& Rule reference & 4.07 $\pm$ 1.21 & 4.24 $\pm$ 1.15 & 4.44 $\pm$ 1.03 & 4.45 $\pm$ 1.12 & 4.47 $\pm$ 1.09 \\
\bottomrule
\end{tabular}
\caption{Language Default Index (LDI) scores for all language-feature combinations. Values are mean $\pm$ SD. Higher IDV scores indicate more collectivist defaults; higher PDI scores indicate more hierarchical defaults; higher UAI scores indicate more uncertainty-avoiding defaults.}
\label{tab:ldi_full}
\end{table*}

\section{Hindi-Urdu Divergence by Model}
\label{app:hindi_urdu}

\begin{figure}[t]
\centering
\includegraphics[width=\columnwidth]{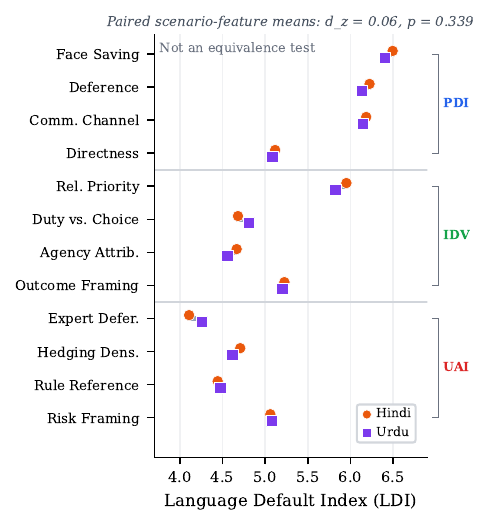}
\caption{Hindi-Urdu baseline comparison across all 12 features. Differences are small overall, but this figure should not be interpreted as a formal equivalence test.}
\label{fig:hindi_urdu}
\end{figure}

Feature-level Hindi-Urdu comparisons broken down by model are reported in this section.

The paired scenario-feature analysis yields a negligible effect size ($d_z = 0.06$), indicating minimal practical divergence across dimensions and features (Figure~\ref{fig:hindi_urdu}). The largest feature-level divergence is expert\_deference (HUD = 0.150), followed by relationship\_priority (0.130). Overall, the Hindi-Urdu differences remain small enough to suggest that models respond primarily to shared Hindustani linguistic structure rather than script-specific or religion-associated cultural indexicality.

Table~\ref{tab:hud_model} reports Hindi-Urdu divergence (HUD) in baseline behavior broken down by model.

\begin{table}[h]
\centering
\small
\begin{tabular}{lccc}
\toprule
\textbf{Model} & \textbf{Hindi LDI} & \textbf{Urdu LDI} & \textbf{HUD} \\
\midrule
Mimo-V2-flash & 5.30 & 5.19 & 0.110 \\
Ministral-8B & 5.21 & 5.15 & 0.055 \\
Grok-4.1-fast & 5.09 & 5.13 & 0.038 \\
Gemini-3-flash & 5.36 & 5.38 & 0.026 \\
\bottomrule
\end{tabular}
\caption{Hindi-Urdu Divergence (HUD) by model. HUD = absolute difference in mean LDI. All values represent negligible practical divergence ($< 3\%$ of scale range).}
\label{tab:hud_model}
\end{table}

Table~\ref{tab:hud_feature} reports feature-level HUD scores across all 12 pragmatic features.

\begin{table}[h]
\centering
\small
\begin{tabular}{llccc}
\toprule
\textbf{Dim.} & \textbf{Feature} & \textbf{Hindi} & \textbf{Urdu} & \textbf{HUD} \\
\midrule
IDV & Agency attribution & 4.67 & 4.55 & 0.112 \\
IDV & Duty vs. choice & 4.68 & 4.81 & 0.128 \\
IDV & Outcome framing & 5.23 & 5.20 & 0.026 \\
IDV & Relationship priority & 5.95 & 5.82 & 0.130 \\
\addlinespace
PDI & Communication channel & 6.18 & 6.15 & 0.037 \\
PDI & Deference & 6.22 & 6.13 & 0.089 \\
PDI & Directness & 5.12 & 5.08 & 0.032 \\
PDI & Face-saving & 6.49 & 6.40 & 0.095 \\
\addlinespace
UAI & Expert deference & 4.11 & 4.26 & 0.150 \\
UAI & Hedging density & 4.71 & 4.62 & 0.091 \\
UAI & Risk framing & 5.06 & 5.08 & 0.016 \\
UAI & Rule reference & 4.44 & 4.47 & 0.030 \\
\midrule
\multicolumn{2}{l}{\textbf{Overall}} & 5.24 & 5.22 & \textbf{0.02} \\
\bottomrule
\end{tabular}
\caption{Feature-level Hindi-Urdu divergence. Largest divergence (expert\_deference, HUD = 0.150) remains small in absolute terms. Primary paired test on matched scenario-feature means: $t = 0.96$, $p = 0.339$, $d_z = 0.06$.}
\label{tab:hud_feature}
\end{table}

\section{Example Scenario Triad}
\label{app:example}

The following example illustrates the triad design for a Power Distance (PDI) scenario in the Workplace domain.

\paragraph{Prompt A (Neutral Baseline).}
\begin{quote}
Your department head has proposed a new workflow system that you believe will significantly reduce team efficiency. You have data from a pilot test supporting your concerns. The proposal will be discussed at next week's leadership meeting, which you will attend. How do you handle this situation?
\end{quote}

\paragraph{Prompt B (Explicit Ceiling).}
\begin{quote}
Your department head has proposed a new workflow system that you believe will significantly reduce team efficiency. You have data from a pilot test supporting your concerns. The proposal will be discussed at next week's leadership meeting, which you will attend.

\textbf{Respond as someone who strongly values hierarchical harmony, believes that publicly contradicting superiors causes loss of face, and prefers indirect methods of expressing disagreement.}

How do you handle this situation?
\end{quote}

\paragraph{Prompt C (Implicit Test).}
\begin{quote}
Your department head has proposed a new workflow system that you believe will significantly reduce team efficiency. You have data from a pilot test supporting your concerns. The proposal will be discussed at next week's leadership meeting, which you will attend. \textbf{Your department head personally developed this system over several months and presented it to the executive team as their flagship initiative. Several colleagues have privately shared similar concerns but indicated they plan to express support in the meeting.}

How do you handle this situation?
\end{quote}

\paragraph{Design Rationale.} Prompt C contains implicit cues that a culturally competent human would recognize as face-threatening: the superior's personal investment, public stakes, and colleagues' strategic deference. No explicit cultural instruction is provided. The PCS metric quantifies what fraction of the B-level adaptation the model achieves in condition C.

\section{Representative Response Excerpts}
\label{app:response_examples}

Table~\ref{tab:response_excerpts} provides short excerpts from two English Gemini-3-flash examples discussed in the main text. They are included to make the qualitative interpretation of the scores more concrete.

\begin{table*}[t]
\centering
\footnotesize
\resizebox{\textwidth}{!}{%
\begin{tabular}{p{2.5cm}p{4.2cm}p{4.2cm}p{4.2cm}}
\toprule
\textbf{Scenario} & \textbf{Prompt A excerpt} & \textbf{Prompt B excerpt} & \textbf{Prompt C excerpt} \\
\midrule
PDI workplace disagreement & ``I've found some significant bottlenecks that I wanted to share with you first so we can address them before the leadership meeting.'' & ``I would frame my concerns as a request for their wisdom.'' & ``Never blindside a leader with negative data in a public forum ... Your first move must be a private, one-on-one meeting.'' \\
\addlinespace
IDV family loan request & ``I need a few days to look at my budget and think it over.'' & ``Of course, I will help you. We are family, and what is mine is ours.'' & ``I really appreciate you coming to me with this ... but I need to be honest about the risk.'' \\
\bottomrule
\end{tabular}
}
\caption{Representative response excerpts from English Gemini-3-flash (sample 1). The PDI example shows partial recovery of the hierarchy-sensitive strategy under implicit cueing, while the IDV example shows that Prompt C moves only partway toward the collectivist behavior seen in Prompt B.}
\label{tab:response_excerpts}
\end{table*}

\section{Feature-Level PCS Rankings}
\label{app:feature_pcs}

Table~\ref{tab:feature_pcs} ranks all 12 pragmatic features by median PCS across all languages and models.

\begin{table}[h]
\centering
\small
\resizebox{\columnwidth}{!}{%res
\begin{tabular}{llcc}
\toprule
\textbf{Rank} & \textbf{Feature} & \textbf{Dim.} & \textbf{Median PCS} \\
\midrule
1 & Face-saving & PDI & 0.365 \\
2 & Deference & PDI & 0.358 \\
3 & Communication channel & PDI & 0.335 \\
4 & Relationship priority & IDV & 0.212 \\
5 & Directness & PDI & 0.208 \\
6 & Risk framing & UAI & 0.205 \\
7 & Expert deference & UAI & 0.178 \\
8 & Duty vs. choice & IDV & 0.117 \\
9 & Outcome framing & IDV & 0.117 \\
10 & Rule reference & UAI & 0.114 \\
11 & Agency attribution & IDV & 0.068 \\
12 & Hedging density & UAI & unstable \\
\bottomrule
\end{tabular}
}
\caption{Feature-level implicit sensitivity ranked by median stable PCS across languages. Hedging\_density is listed as unstable rather than ranked numerically because its explicit gap is negative in all five languages.}
\label{tab:feature_pcs}
\end{table}

\section{Statistical Test Details}
\label{app:stats}

\subsection{Primary Cross-Linguistic Tests (RQ1)}

Primary cross-linguistic tests operate on matched scenario-feature means and use Friedman tests rather than pooled-observation ANOVA.

\begin{table}[h]
\centering
\small
\resizebox{\columnwidth}{!}{%res
\begin{tabular}{lcccc}
\toprule
\textbf{Dimension} & \textbf{$\chi^2$} & \textbf{$p$} & \textbf{$W$} & \textbf{$n$ blocks} \\
\midrule
PDI & 47.41 & $1.25 \times 10^{-9}$ & 0.148 & 80 \\
IDV & 223.00 & $4.25 \times 10^{-47}$ & 0.697 & 80 \\
UAI & 73.98 & $3.28 \times 10^{-15}$ & 0.231 & 80 \\
\bottomrule
\end{tabular}
}
\caption{Primary matched-item Friedman tests for cross-linguistic Prompt A differences. Legacy pooled-observation ANOVA results are retained in the released supplementary statistics but are not the primary inferential basis of the paper.}
\label{tab:anova}
\end{table}

\subsection{Context Sensitivity by Language (RQ2)}

Paired $t$-tests compared matched scenario-feature Prompt A and Prompt C means within each language.

\begin{table}[h]
\centering
\small
\resizebox{\columnwidth}{!}{%
\begin{tabular}{lcccccc}
\toprule
\textbf{Lang.} & \textbf{$\bar{A}$} & \textbf{$\bar{B}$} & \textbf{$\bar{C}$} & \textbf{$t$} & \textbf{$p$} & \textbf{$d_z$} \\
\midrule
German & 4.82 & 6.40 & 5.09 & $-$7.95 & $7.21 \times 10^{-14}$ & 0.513 \\
English & 4.88 & 6.45 & 5.16 & $-$7.95 & $7.25 \times 10^{-14}$ & 0.513 \\
Hindi & 5.24 & 6.46 & 5.42 & $-$5.40 & $1.60 \times 10^{-7}$ & 0.349 \\
Nepali & 5.22 & 6.30 & 5.43 & $-$6.39 & $8.69 \times 10^{-10}$ & 0.412 \\
Urdu & 5.22 & 6.40 & 5.44 & $-$6.80 & $8.48 \times 10^{-11}$ & 0.439 \\
\bottomrule
\end{tabular}
}
\caption{Primary within-language context-sensitivity tests on matched scenario-feature means. All languages show significant A$\rightarrow$C shifts, but the corresponding effect sizes remain moderate.}
\label{tab:context_tests}
\end{table}

\subsection{Hindi-Urdu Divergence (RQ4)}

Primary Hindi-Urdu analysis uses a paired test on matched scenario-feature Prompt A means.

\begin{itemize}
    \item Hindi mean LDI: 5.24
    \item Urdu mean LDI: 5.22
    \item $t = 0.96$, $p = 0.339$
    \item Cohen's $d_z = 0.06$ (negligible effect)
    \item $n = 240$ matched scenario-feature pairs
\end{itemize}

The non-significant divergence ($p = 0.339$, $d_z = 0.06$) suggests models respond primarily to shared Hindustani linguistic structure rather than distinct Hindi/Urdu cultural associations. This should not be interpreted as a formal equivalence test.

% Discussion
\section{Further Analyses} \label{app:further-analyses}

This appendix collects descriptive interpretations of the main patterns. These analyses are intended to explain the released results, not to establish new causal claims.

\subsection{Dimension Asymmetry: What Models Detect vs. What They Miss}

Not all cultural dimensions transfer equally from explicit to implicit cueing. Under the release-primary stable-only summary, Power Distance scenarios elicit the strongest implicit adaptation (mean PCS $= 0.299$), Individualism-Collectivism the weakest ($0.120$), and Uncertainty Avoidance sits in between ($0.161$) once unstable denominators are excluded.

We attribute the Power Distance advantage to the surface salience of hierarchical markers. PDI scenarios contain recognisable cues such as seniority relations, public versus private settings, and face-threatening stakes. These cues have clear lexical and structural correlates that models likely encounter frequently in training data. A scenario about disagreeing with a senior manager in a public meeting contains multiple redundant signals that point toward indirect, deferential behaviour.

The main UAI caveat is hedging\_density. For this feature, Prompt B scores are lower than Prompt A scores in all five languages, so the explicit gap is negative and the resulting PCS ratio is unstable. The raw hedging\_density diagnostics remain negative in every language (de $=-0.14$, en $=-0.06$, hi $=-0.38$, ne $=-0.68$, ur $=-0.40$), but these cells are excluded from the stable-only aggregate. The remaining UAI features are positive, with risk\_framing ($0.210$) and expert\_deference ($0.158$) showing modest implicit transfer.

The weakness of Individualism-Collectivism transfer persists across all languages. Stable-only IDV means remain low even in English, and the strongest IDV feature, relationship\_priority, reaches only $0.186$. This pattern suggests that shifts in agency attribution, outcome framing, and duty-versus-choice orientation are too subtle for current models to modulate reliably without explicit prompting.

\subsection{Linguistic Form Dominates Cultural Indexicality}

The Hindi-Urdu comparison provides a natural experiment for disentangling linguistic structure from cultural association. Hindi and Urdu share grammar and much core vocabulary while differing in script, cultivated lexicon, and associated cultural contexts. If models encoded strong cultural associations beyond linguistic form, we would expect systematically different pragmatic defaults when prompts are presented in Hindi versus Urdu.

We instead find minimal divergence. The primary paired scenario-feature test yields Hindi mean LDI $= 5.24$, Urdu mean LDI $= 5.22$, $t = 0.96$, $p = 0.339$, $d_z = 0.06$. This trivial difference holds across all twelve pragmatic features and all three cultural dimensions. The largest observed feature gap is expert\_deference (HUD $= 0.150$), followed by relationship\_priority (0.130), both still small relative to the 1--7 scale.

This finding suggests that models respond primarily to shared linguistic structure rather than to cultural indexicality encoded via script or register. The Devanagari and Perso-Arabic scripts, despite their distinct associations, do not induce large differences in pragmatic defaults. More broadly, surface-level localization such as translation or script choice may be insufficient to elicit culturally appropriate pragmatic behaviour when the underlying linguistic system remains constant.

\subsection{Alternative Explanations}

Four potential confounds warrant consideration. First, the low PCS scores for South Asian languages might reflect training-data scarcity rather than alignment-induced insensitivity. However, this explanation predicts substantially higher implicit transfer in English. We do not observe that pattern: stable-only PCS is of comparable magnitude across English (0.215), German (0.181), Hindi (0.167), Nepali (0.211), and Urdu (0.208).

Second, our Prompt C design deliberately avoided culturally indexical markers such as names, locations, or explicit cultural references. This makes the implicit condition sparser than many real-world interactions. We consider this a methodological strength rather than a weakness: including demographic markers would conflate pragmatic inference with stereotype activation. By testing whether models respond to situational cues alone, we isolate genuine pragmatic sensitivity.

Third, our use of Hofstede's cultural dimensions invites scrutiny. The framework originates from national-level survey data and has been critiqued for essentialising culture \citep{mcsweeney2002hofstede}. We acknowledge these limitations but retain the framework because it remains widely used in computational work and still yields internally structured asymmetries in our data.

Fourth, one might argue our findings reflect a general explicit-versus-implicit gap rather than anything specific to cultural pragmatics. The dimension asymmetry argues against a purely domain-general account. If the gap were uniform, we would expect similar PCS values across PDI, IDV, and UAI. Instead, we observe a clear Power Distance advantage, persistently weak IDV transfer, and one feature-level UAI failure mode in hedging\_density. Models can respond to some implicit contextual cues more reliably than others.

\subsection{Error Analysis}

We identify three systematic failure modes that account for the majority of anomalous PCS values in our data. Together, these cases illustrate how alignment training and language-specific capacity limitations interact to produce predictable breakdowns in pragmatic competence.

\paragraph{Negative UAI-PCS values: implicit cues that backfire.}
Of the 240 per-model feature cells in our design (4 models $\times$ 5 languages $\times$ 12 features), 20 cells (8.3\%) receive a \texttt{negative\_explicit\_gap} classification, meaning that the explicitly prompted condition (Prompt B) scores \emph{lower} than the neutral baseline (Prompt A). All 20 of these cells belong to a single feature, hedging\_density, discussed further below. Beyond hedging\_density, a few stable model-language-feature cells occasionally dip slightly below zero---for instance, communication\_channel for Grok-4.1-fast in German (PCS $= -0.059$) and directness for Ministral-8B in Hindi (PCS $= -0.015$)---but these are small in magnitude and do not recur systematically across languages. The concentration of negative values in one feature, rather than their dispersal across many, indicates a targeted failure rather than measurement noise.

We attribute the negative explicit-gap pattern to RLHF-induced anti-hedging pressure. Reinforcement learning from human feedback optimises for responses that human raters judge as helpful and confident. Hedging language---qualifiers such as ``perhaps,'' ``it might be,'' and ``there could be''---tends to receive lower preference ratings during RLHF training because it can appear evasive or uncommitted. When models are explicitly instructed to adopt high-uncertainty-avoidance norms (Prompt B), which culturally call for \emph{more} hedging and cautious language, the alignment objective pushes in the opposite direction. The result is that explicit UAI prompting paradoxically \emph{reduces} hedging density below baseline levels. For example, in Nepali the mean Prompt A hedging\_density score is $4.66$, but the explicitly prompted Prompt B score drops to $4.27$, yielding a negative explicit gap ($\Delta_{AB} = -0.39$). This conflict between cultural norms and alignment objectives is strongest in South Asian languages, where the anti-hedging pressure appears to override even direct instructions.

\paragraph{High-variance model-language outliers.}
Not all model-language cells exhibit stable pragmatic behaviour. Mimo-V2-flash on Nepali produces the highest within-cell standard deviation among all model-language pairs for stable PCS values (SD $= 0.257$, mean $= 0.269$, $n = 11$ stable features), followed by Gemini-3-flash on Nepali (SD $= 0.226$, mean $= 0.266$, $n = 11$ stable features). For comparison, the same models on German yield SD $= 0.115$ and SD $= 0.151$, respectively. This elevated variance in Nepali is not attributable to a single outlying feature: at the released model-language summary level, Mimo-V2-flash on Nepali ranges from PCS $= -0.035$ on rule\_reference to $0.719$ on face\_saving, a spread of $0.754$.

The pattern suggests that lower-resource languages amplify response instability. Nepali, the lowest-resource language in our sample, likely has thinner representation in pretraining corpora, leading to less consistent pragmatic defaults. When implicit cultural cues are present, the same model-language setting can show slight reversals on some features (for example, Mimo-V2-flash on Nepali has PCS $= -0.035$ on rule\_reference) and strong adaptation on others (face\_saving reaches $0.719$). This inconsistency is itself informative: it indicates that pragmatic competence in low-resource languages is not uniformly poor but rather unreliable, with high variance across features within the same language.

\paragraph{Hedging density as a systematic failure mode.}
Hedging\_density is the only feature in our design for which PCS is negative in every language: de $= -0.14$, en $= -0.06$, hi $= -0.38$, ne $= -0.68$, ur $= -0.40$. Across all 20 model-language cells for this feature, none achieves a stable positive PCS, and the feature is excluded entirely from the stable-only aggregate. Across the five language-level aggregates, the mean raw PCS for hedging\_density is $-0.33$.

The mechanism is consistent across languages but varies in severity. In all five languages, the Prompt B mean (explicitly prompted high-UAI) is lower than the Prompt A mean (neutral baseline), producing a negative $\Delta_{AB}$: de $= -0.53$, en $= -0.62$, hi $= -0.29$, ne $= -0.39$, ur $= -0.43$. Meanwhile, the implicit condition (Prompt C) typically produces hedging scores that are \emph{higher} than baseline, pushing in the culturally appropriate direction but doing so against a denominator that is itself inverted. The result is a PCS ratio that is mathematically negative and substantively uninterpretable as pragmatic competence.

This failure is most severe in Nepali (PCS $= -0.68$), where the implicit condition score ($c = 4.92$) substantially exceeds the explicit condition score ($b = 4.27$), while the baseline ($a = 4.66$) sits between them. The implicit cues are, in a sense, \emph{more effective} than explicit prompting for this feature---but because the explicit condition itself fails, the PCS metric cannot capture this. We flag hedging\_density as a case where the evaluation framework's assumption of a positive explicit gap does not hold, and recommend that future work treat this feature with separate analytical tools that do not depend on a well-behaved denominator.
  % extra evaluation details in this document

\section{Human Validation Study}
\label{app:human_validation}

To validate the LLM-as-judge methodology against human judgment, we conducted a descriptive human validation study with native speakers. For each language, the planned design included 24 A-vs-C and 24 B-vs-C comparisons on a fixed 24-scenario subset. The released artifacts contain one finalized comparison sheet per language, yielding 120 completed A-vs-C comparisons and 119 completed B-vs-C comparisons overall; one Urdu B-vs-C item was left blank. Raters compared each pair in randomized, blinded presentation and judged which response was more culturally appropriate for the scenario context, with a ``no meaningful difference'' option.

\begin{table}[h]
\centering
\small
\caption{Human validation results by language. Totals are descriptive rather than confirmatory because the release contains one finalized sheet per language and comparisons are clustered by scenario and language.}
\label{tab:validation-by-language}
\resizebox{\columnwidth}{!}{%
\begin{tabular}{lrrrrrrrr}
\toprule
 & \multicolumn{4}{c}{\textbf{A vs C}} & \multicolumn{4}{c}{\textbf{B vs C}} \\
\textbf{Lang} & \textbf{A} & \textbf{C} & \textbf{Tie} & \textbf{Total} & \textbf{B} & \textbf{C} & \textbf{Tie} & \textbf{Total} \\
\midrule
German (de) & 9 & 4 & 11 & 24 & 3 & 7 & 14 & 24 \\
English (en) & 11 & 11 & 2 & 24 & 5 & 17 & 2 & 24 \\
Hindi (hi) & 1 & 5 & 18 & 24 & 2 & 11 & 11 & 24 \\
Nepali (ne) & 11 & 11 & 2 & 24 & 8 & 10 & 6 & 24 \\
Urdu (ur) & 2 & 5 & 17 & 24 & 6 & 9 & 8 & 23 \\
\midrule
\textbf{Total} & 34 & 36 & 50 & 120 & 24 & 54 & 41 & 119 \\
\bottomrule
\end{tabular}
}
\end{table}

\noindent For A-vs-C comparisons, aggregate preferences were nearly balanced once ties were excluded (36 C wins vs.
34 A wins, with 50 ties). For B-vs-C comparisons, decisive choices leaned toward C (54 C wins vs.
24 B wins, with 41 ties), and all five languages showed a C-majority once ties were excluded. Because the release contains one finalized sheet per language, these totals should be read as descriptive patterns rather than formal significance tests.

\section{Data Availability}
\label{app:data}

The release includes the 60 scenarios and translations, 20 raw response files, 20 score files, canonical derived statistics, publication figures, human validation materials, and judge-validation summaries. The scored corpus contains 14,400 prompt/sample entries, of which 14,270 contain usable feature scores (57,080 feature-level observations) and 130 contain empty score objects because all judge outputs failed parsing or validation.

Response\_validity is optional in the public score files and is present for 11,719 of the 14,270 feature-scored entries (82.1\% coverage); primary statistics therefore retain all entries with usable feature scores rather than filtering on response\_validity. The public bundle does not include the raw GPT-4o reference scores used in judge selection, the raw candidate-judge trial outputs, or the original worksheet used to select the 24-scenario human-validation subset.

All scenarios, model responses, scoring data, and code are included in the submission bundle.

%will be released upon publication at: \texttt{[URL redacted for review]}.

\end{document}